\documentclass[final,3p,times,twocolumn,authoryear]{elsarticle}
\usepackage{xurl}  

\usepackage{subcaption}

\usepackage{algorithm2e}
\RestyleAlgo{ruled}

\usepackage{breqn}

\usepackage{comment}

\usepackage{amssymb}

\usepackage{multirow}

\usepackage{tabularx}

\usepackage[dvipsnames]{xcolor}

\newcommand{\resultsImgSize}{0.25\textwidth}
\newcommand{\inputsImgSize}{0.25\textwidth}
\newcommand{\probabilitiesImgSize}{0.45\textwidth}
\newcommand{\intentionImgSize}{0.4\textwidth}

\usepackage{cleveref}
\crefname{figure}{Fig.}{Figs.}
\creflabelformat{figure}{(#2#1#3)} % with brackets
\creflabelformat{figure}{#2#1#3} % without brackets

\journal{Ocean Engineering}

\begin{document}

\begin{frontmatter}


\title{Design and Validation of an Intention-Aware Probabilistic Framework for Trajectory Prediction: Integrating COLREGS, Grounding Hazards, and Planned Routes\tnoteref{label1}}
\tnotetext[label1]{The research leading to these results has received funding from the European Union’s Horizon 2020 research and innovation program under the Marie Skłodowska-Curie grant agreement No 955.768 (MSCA-ETN AUTOBarge). This publication reflects only the authors’ view, exempting the European Union from any liability. Project website: http://etn-autobarge.eu/.}
\author{Dhanika Mahipala\corref{cor1}\fnref{aff1}}
\ead{dhanika.mahipala@km.kongsberg.com}
\author{Trym Tengesdal\fnref{aff2}}
\ead{trym.tengesdal@ntnu.no}
\author{Børge Rokseth\fnref{aff2}}
\ead{borge.rokseth@ntnu.no}
\author{Tor Arne Johansen\corref{cor1}\fnref{aff2}}
\ead{tor.arne.johansen@ntnu.no}
\cortext[cor1]{Corresponding authors}
\affiliation[aff1]{organization={Cybernetics and Simulation - ISOL},
           addressline={Kongsberg Maritime AS}, 
           country={Norway}}
\affiliation[aff2]{organization={Department of Engineering Cybernetics},
           addressline={Norwegian University of Science and Technology}, 
           country={Norway}}

% \title{Design and Validation of an Intention-Aware Probabilistic Framework for Trajectory Prediction: Integrating COLREGS, Grounding Hazards, and Planned Routes} %% Article title

% %% use optional labels to link authors explicitly to addresses:
% \author[label1,label2,label2,label2]{Dhanika Mahipala,Trym Tengesdal,Børge Rokseth,Tor Arne Johansen}
% \affiliation[label1]{organization={Kongsberg Maritime AS},
%             country={Norway}}

% \affiliation[label2]{organization={Norwegian University of Science and Technology},
%             country={Norway}}

% \author{} %% Author name

% \affiliation{organization={},%Department and Organization

%% Abstract
\begin{abstract}
Collision avoidance capability is an essential component in an autonomous vessel navigation system. To this end, an accurate prediction of dynamic obstacle trajectories is vital. Traditional approaches to trajectory prediction face limitations in generalizability and often fail to account for the intentions of other vessels. While recent research has considered incorporating the intentions of dynamic obstacles, these efforts are typically based on the own-ship's interpretation of the situation. The current state-of-the-art in this area is a Dynamic Bayesian Network (DBN) model, which infers target vessel intentions by considering multiple underlying causes and allowing for different interpretations of the situation by different vessels. However, since its inception, there have not been any significant structural improvements to this model. In this paper, we propose enhancing the DBN model by incorporating considerations for grounding hazards and vessel waypoint information. The proposed model is validated using real vessel encounters extracted from historical Automatic Identification System (AIS) data.
\end{abstract}


% \begin{highlights}
% \end{highlights}

%% Keywords

\end{frontmatter}



\section{Introduction} \label{sec:introduction}
An essential function of an autonomous vessel is its capability to avoid collisions. Many collision avoidance algorithms have been proposed over the years \cite{vagale_path_2021,huang_ship_2020,akdag_collaborative_review} where different approaches were considered to fulfill the task. Most of these algorithms require the prediction of the trajectory of surrounding vessels (target vessels), within a predefined time horizon. For example, the Scenario-Based Model Predictive Control algorithm \cite{johansen_ship_2016} proposes a simple kinematic model that assumes a straight-line trajectory (constant velocity and course). 
% Another example is \cite{ship_colav_and_anti_grounding}, where multiple possible trajectories, generated through constant course and speed modifications to the current state of the vessel, are considered. 

Correctly predicting future trajectories of target vessels is crucial for any collision avoidance algorithm. Hence, it is worthwhile to focus separately on this aspect alone. Article \cite{li_ship_2023} summarizes ship trajectory prediction methods, grouping the relevant literature into three main categories, ship model-based, artificial intelligence-based and deep learning-based. Within the context of collision avoidance, the papers such as \cite{xie_ship_2019} from ship model-based methods, \cite{pedrielli_real_2020,zheng_decision-making_2021,chen_identification_2021} from artificial intelligence-based methods, and \cite{yang_ship_2023,zhang_meto-s2s_2023} from deep learning-based methods can be highlighted from the survey. Most of these studies rely on historical Automatic Identification System (AIS) data, which can make it difficult to generalize the findings due to its inherent heterogeneous and multi-modal nature of motion data \cite{nguyen_transformer_2024}. Additionally, they do not explicitly consider the intentions of other ships. 

There is, however, other research where target vessel intentions are explicitly considered to some extent. \cite{cheng_human-like_2019} propose a human-like decision-making model using a Bayesian Belief Network for ship crossings, taking into account ship behavior and maneuverability. Another group of researchers (\cite{cho_intent_2018, cho_intent_2021, cho_intent_2022}) has attempted to model the maneuver intentions of obstacle ships using probabilistic tools.

According to COLREGs, the stand-on vessel is not permitted to maneuver until it is evident that the give-way vessel is not taking proper and timely action. To address this, researchers \cite{du_improving_2020} have proposed a framework to help determine the intentions of the give-way vessel, which is crucial for the stand-on vessel in deciding when to take evasive action. Moreover, \cite{tran_collision_2023-1} proposes a proactive collision avoidance algorithm that utilizes situation-based intention prediction for neighboring ships.

However, the aforementioned research evaluates target vessel intentions based on the own-ship's interpretation of the situation. This can lead to discrepancies—for example, the own-ship and the target ship may have differing assessments of the collision risk scenario. Their interpretations of a safe passing distance or the timeliness of necessary actions to prevent a collision might also vary. Such underlying causes are not accounted for in previous research. The International Regulations for Preventing Collisions at Sea (COLREGs) \cite{imo_convention_1972} themselves are vague and leave room for skippers' experience and good seamanship to influence how situations are interpreted. The state-of-the-art in this area is \cite{rothmund_intention_2022}, which proposes a target vessel intention inference method that considers multiple underlying causes, allowing for different interpretations of the situation by different vessels.

The method proposed in \cite{rothmund_intention_2022} uses a Dynamic Bayesian Network (DBN) model to infer the intentions of a target vessel while taking underlying causes into consideration. A DBN is a subgroup of networks within the domain of Bayesian Belief Networks (BBNs). BBNs are directed acyclic graphs consisting of nodes and arcs that connect them, used to model probabilistic relationships. Each node consists of a discrete set of states, and arcs are defined using Conditional Probability Tables (CPTs), where the probability of each state in a child node is represented as a function of the states of its parent nodes. The CPT of a root node (a node without a parent) defines its prior probability distribution. The prior probability distributions of root nodes are determined based on historical data and/or expert surveys.

Evidence refers to information about the state of a particular node. If this information is uncertain, virtual evidence can be used. Once evidence is set for all known nodes, Bayes' Theorem can be applied to calculate the probabilities of the states of unknown nodes. BBNs can be extended dynamically by incorporating time-dependent nodes, also known as temporal nodes. This special class of BBNs is referred to as DBNs.

Since \cite{rothmund_intention_2022}, a few minor improvements and verification studies have been carried out. Article \cite{rothmund_validation_2023} is one such study, where the intention model proposed in \cite{rothmund_intention_2022} was validated in simulation using AIS data. Another study, \cite{tengesdal_obstacle_2024}, uses the model in a collision avoidance algorithm and validates the outcome in a real-time experimental setting. However, there haven't been any structural modifications to the DBN proposed in \cite{rothmund_intention_2022} to extend its capabilities.

\subsection{Problem Statement and Contribution} \label{subsec:contribution}
There are several shortcomings in the existing DBN model \cite{rothmund_intention_2022} proposed for target vessel intention inference. One such shortcoming is the lack of consideration of grounding hazards within the model. The state of the grounding hazards in the vicinity of the vessel greatly affects decision-making. For example, a target vessel could violate its give-way responsibility towards its starboard side due to static obstacles present on its starboard. Another shortcoming is the inability to consider waypoint information of a target vessel. Depending on the location of the next waypoint, a target vessel may or may not decide to engage in a collision avoidance maneuver with the own-ship, as navigating towards that waypoint could avoid the risk of collision altogether. Hence, this research aims to address the following primary objectives:
\begin{itemize}
    \item Enhancing the capabilities of the existing DBN model by incorporating waypoint information.
    \item Enhancing the capabilities of the existing DBN model by incorporating information on grounding hazards in the vicinity.
    \item Proposing a methodology to extract values for tunable parameters of the model using historical AIS data.
    \item Testing and validation of the improved DBN model through real vessel encounters from historical AIS data.
\end{itemize}

\section{Overview} \label{sec:overview}
The proposed DBN presented in this paper is developed taking the following assumptions into consideration:
\begin{itemize}
    \item \textbf{Assumption 01:} Our methodology extends the capabilities of the model presented in \cite{rothmund_intention_2022}. However, to improve clarity, certain nodes in the original model, such as Situation Started (SS) and Risky Situation (RS), are not included, as they do not necessarily contribute to the core objective of our model. Hence, the model presented in this paper is assumed to be used in situations where there is no uncertainty regarding when the situation starts or whether there is a risk of collision, similar to \cite{rothmund_validation_2023}.
    \item \textbf{Assumption 02:} The next waypoint of each target vessel is known at any given point in time. This is sometimes a reasonable assumption, as it can be achieved through various means, such as using historical AIS data or route exchange methods \cite{tengesdal_risk-based_2020}.
\end{itemize}

Our DBN model (see Section \ref{sec:network_architecture}) needs to be evaluated separately from the point of view of each target vessel in the vicinity. The target vessel under consideration will be referred to as the "reference vessel" from here onwards. Each and every vessel, apart from the reference vessel will be referred to as an "obstacle vessel". Each obstacle vessel will be assigned an index number designated by the variable $i$ in the upcoming sections. The DBN nodes with the subscript $i$ should be added to the model in multiples of the number of vessels under consideration. There are known nodes and unknown nodes in the model. The known nodes represent various measurements that can be determined from the current state of the reference vessel i.e., latitude, longitude, speed over ground (SOG), and course over ground (COG). The unknown nodes are the intention nodes, which represent the intention information of the reference vessel that we want to predict. For example, these may include what the reference vessel considers as ample time or what it considers as a safe distance from another vessel. The relationships among the DBN nodes are structured in a way that allows evaluation of whether a particular combination of measurements and intention node states are compatible. These combinations are determined according to COLREGS rules 7, 8, 11, and 13 through 18.

The process of using the proposed DBN for real-time intention inference and trajectory evaluation can be divided into several steps. First the DBN should be initialized by clearing all evidence and setting the number of time slices to one. Next, within a loop at every single time step of the simulation,
\begin{itemize}
    \item \textbf{Step 01}: Calculate the states for the measurement nodes using the current state of the reference vessel and the obstacle vessels.
    \item \textbf{Step 02}: Set evidence for the measurement nodes using the findings from \textbf{Step 01}.
    \item \textbf{Step 03}: Set the evidence of node variable $C$ (see Eq. \eqref{eq:c}) to $true$\footnote{More information on each variable can be found in Section \ref{sec:network_architecture}.}.
    \item \textbf{Step 0}4: Update the beliefs of the network and find the probability distributions for the intention nodes.
    \item \textbf{Step 05}: Add a temporary time slice to the DBN.
    \item \textbf{Step 06}: Generate candidate trajectories.
    \item \textbf{Step 07}: For each candidate trajectory:
    \begin{itemize}
        \item Calculate the states of the measurement nodes and set evidence for the measurement nodes in the temporary time slice added in \textbf{Step 05}. Some measurement nodes are evaluated differently compared to how it is done in \textbf{Step 01}. More information on these slight changes can be found in Section \ref{subsubsec:measurements_candidate_traj}.
        \item Use the probability  distributions of intention nodes found in \textbf{Step 04} to set virtual evidence of the intention nodes in the temporary time slice added in \textbf{Step 05}.
        \item Update the beliefs of the network and find the probability of $true$ state in the node $C$. This is the probability of compatibility between that specific candidate trajectory and the intentions.
    \end{itemize}
    \item \textbf{Step 08}: Delete the temporary time slice added to the network in \textbf{Step 05}.
    \item \textbf{Step 09}: Add a new time slice to the network which will be used for \textbf{Step 01} through \textbf{Step 04} in the next time step of the simulation.
\end{itemize}
Adding a new time slice to the DBN at every time step of the simulation (\textbf{Step 09}) can increase the computational burden exponentially. Hence, as proposed in \cite{tengesdal_obstacle_2024}, a condition can be added to decide whether to add a new time slice or not. For example, a new time slice can be added only when the previous time slice was created more than $\Delta_{ts,max}$ ago, or if the previous time slice was added no less than $\Delta_{ts,min}$ ago and any ship in the encounter has changed its course by more than $\Theta^\circ$ or speed by more than $\Upsilon$ m/s. If the condition is not met, \textbf{Steps 01} through \textbf{04} will be continuously performed on the last time slice of the DBN.

\section{Bayesian Network Structure} \label{sec:network_architecture}
This section describes the DBN structure of the proposed model using a bottom-up approach for reproducibility. First, root nodes—i.e., nodes without parent nodes that represent variables independent within a time slice—are introduced. Then, intermediate and leaf nodes are presented in a replicable order. Fig. 1 shows the DBN's structure for a single ship encounter, with each node detailed in the following sub-sections.
\begin{figure*}[htbp]
    \centering
    \includegraphics[width=0.8\textwidth]{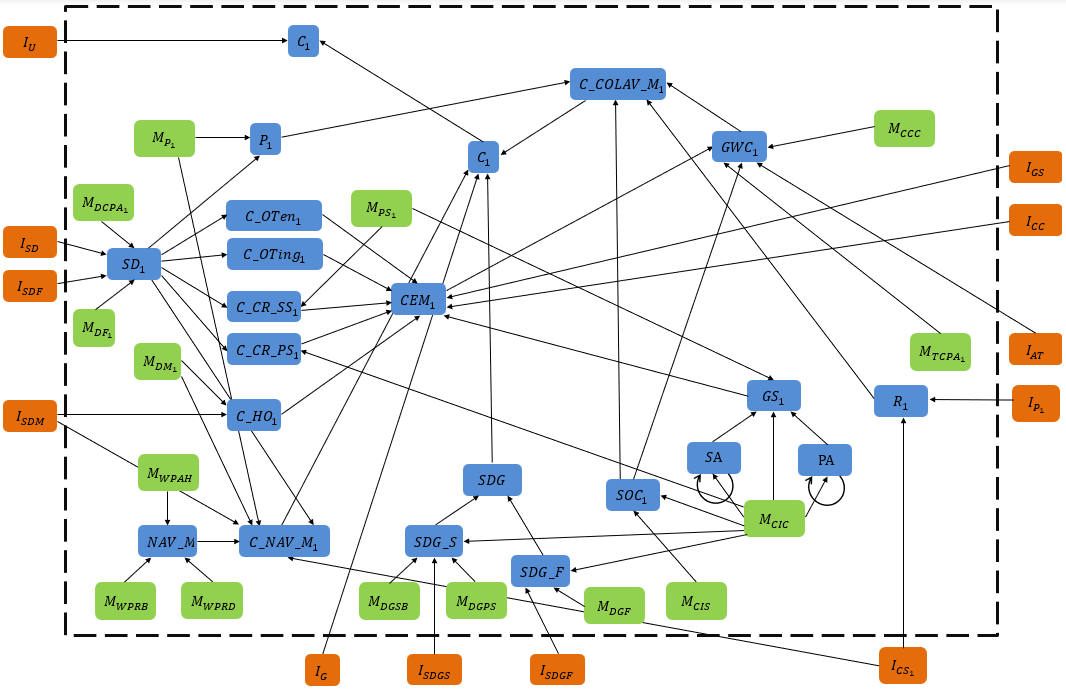}
    \caption{Graphical representation of the topography of the proposed DBN for a single ship encounter. Subscript 1 indicates that this model considers the relation between the reference ship and the ship with index 1. All nodes within the box are time-dependent and are replicated at each time step. Circular arrows represent connections between consecutive time steps.}
    \label{fig:dbn}
\end{figure*}
The DBN includes three node types within the context of the proposed network: intention, measurement, and model variables. Intentions, assumed constant during an encounter, are time-independent nodes, while measurement and model variables are time-dependent. 

\subsection{Intention nodes} \label{subsec:intention_nodes}
Most of the intention nodes in the proposed network are summarized in \cite{rothmund_intention_2022} and \cite{tengesdal_obstacle_2024}. For convenience, all intention nodes including those introduced in this paper are listed in Table \ref{tab:intention_nodes}. There are two main types of intention nodes: binary nodes and real-valued nodes. Binary nodes have two states, "true" or "false," while real-valued nodes can have multiple states, each representing a value range specific to the variable's context.
% \begin{table}[htb]
% \centering
% \caption{Intention nodes}
% \begin{tabularx}{\columnwidth}{|l|X|l|}
% \hline
%         Symbol  & Description & States \\ \hline
%         $I_{G}$  & Whether the reference ship intends to sail towards a grounding hazard. & Binary \\ 
%         $I_{SDGS}$  & What the reference ship considers a safe distance from either sides of the vessel to ground. & Real valued \\ 
%         $I_{SDGF}$  & What the reference ship considers a safe distance from the front of the vessel to ground. & Real valued \\ 
% \hline
% \end{tabularx}
% \label{tab:intention_nodes}
% \end{table}
\begin{table*}[htb]
\centering
\caption{Intention nodes}
\begin{tabularx}{\textwidth}{|l|X|l|}
\hline
        Symbol  & Description & States \\ \hline
        $I_{AT}$  & What time until CPA the reference ship considers ample time. & Real valued \\ 
        $I_{CC}$  & Whether the reference ship intends to be COLREGS-compliant when performing evasive maneuvers. & Binary \\ 
        $I_{GS}$  & Whether the reference ship acts according to good seamanship. & Binary \\ 
        $I_{P_i}$  & Whether the reference ship acts as if it has a lower or higher priority towards ship $i$. & "higher"/ "similar"/ "lower" \\ 
        $I_{CS_i}$  & What COLREGS situation the reference ship thinks it has towards ship $i$. & "OT\_ing"/ "OT\_en"/ "HO"/ "CR\_PS"/ "CR\_SS" \\ 
        $I_{G}$  & Whether the reference ship intends to sail towards a grounding hazard. & Binary \\ 
        $I_{SDGS}$  & What the reference ship considers a safe distance from either sides of the vessel to ground. & Real valued \\ 
        $I_{SDGF}$  & What the reference ship considers a safe distance from the front of the vessel to ground. & Real valued \\ 
        $I_{SD}$  & What the reference ship considers a safe distance at CPA. & Real valued \\ 
        $I_{SDF}$  & How far in front of a ship the reference ship considers a crossing as safe. & Real valued \\ 
        $I_{SDM}$  & What the reference ship considers a safe distance at CPA to the current midpoint. & Real valued \\ 
        $I_{U}$  & Whether the reference ship acts in an unmodeled way. & Binary \\ 
\hline
\end{tabularx}
\label{tab:intention_nodes}
\end{table*}

\subsection{Measurement nodes} \label{subsec:measurement_nodes}
Measurement nodes that are in the intention model proposed in this paper are summarized in Table \ref{tab:measurement_nodes}. The values for the measurement nodes are set as evidence during run-time. Hence, the prior probability distribution must be a uniform distribution. Since the real-valued nodes in the network are either time or distance measurements, lower limit of the distribution is equal to 0 for every node while the upper limit can be determined depending on the expected range for each measurement. The upper limit value of each real-valued node signify the saturation point and any measurement larger than that value would be considered a measurement from the state furthest from the initial state that includes the value zero.
\begin{table*}[htbp]
\centering
\caption{Measurement nodes}
\begin{tabularx}{\textwidth}{|l|X|l|}
\hline
        Symbol  & Description & States \\ \hline
        $\mathcal{M}_{DCPA_i}$  & Distance between reference ship and ship $i$ at CPA assuming both will keep their current course and speed. & Real valued \\ 
        $\mathcal{M}_{DF_i}$  & How far the reference ship crosses in front on ship $i$ assuming both keep their current course and speed. This value is set to $\infty$ if the ship does not cross in front of ship $i$. & Real valued \\ 
        $\mathcal{M}_{DM_i}$  & Distance at CPA to the current midpoint between the reference ship and ship $i$, assuming constant course and speed for the reference ship. & Real valued \\ 
        $\mathcal{M}_{P_i}$  & Whether reference ship has passed ship $i$. & Binary \\ 
        $\mathcal{M}_{PS_i}$  & Whether the reference ship will pass with ship $i$ on its port or starboard side. & "starboard"/ "port" \\ 
        $\mathcal{M}_{MPS_i}$  & Whether the reference ship will pass the current midpoint between itself and ship $i$ on its port or starboard side. & "starboard"/ "port" \\ 
        $\mathcal{M}_{TCPA_i}$  & Time until CPA between reference ships and ship $i$ assuming both will keep their current course and speed. & Real valued \\ 
        $\mathcal{M}_{CIC}$  & The change in course of the reference ship since the initial course at the start of the situation. & "starboard"/ "port"/ "straight" \\ 
        $\mathcal{M}_{CIS}$  & The change in speed of the reference ship since the initial speed at the start of the situation. & "higher"/ "lower"/ "none" \\ 
        $\mathcal{M}_{CCC}$  & Whether the reference ship is currently changing course. & Binary \\ 
        $\mathcal{M}_{CS_i}$  & Current COLREGS situation reference ship has towards ship $i$. & "OT_ing"/ "OT_en"/ "HO"/ "CR_PS"/ "CR_SS" \\
        $\mathcal{M}_{DGSB}$  & Current distance to the closest point of the ground hazard on the starboard side of the vessel. & Real valued \\
        $\mathcal{M}_{DGPS}$  & Current distance to the closest point of the ground hazard on the port side of the vessel. & Real valued \\
        $\mathcal{M}_{DGF}$  & Current distance to the closest point of the ground hazard on the front of the vessel. & Real valued \\
        $\mathcal{M}_{WPRB}$  & The relative bearing difference from the vessel to the next way-point. & \{decreasing, increasing, neither\} \\
        $\mathcal{M}_{WPRD}$  & The distance difference from the vessel to the next way-point. & \{decreasing, increasing, neither\} \\
        $\mathcal{M}_{WPAH}$  & Whether the reference ship is headed towards the next way-point. & \{decreasing, increasing, neither\} \\
\hline
\end{tabularx}
\label{tab:measurement_nodes}
\end{table*}

\subsection{Model nodes} \label{model_nodes}
The rest of the nodes in the network are called model nodes, which include both intermediate and leaf nodes. All model nodes are binary, having two states: 'true' and 'false'. The following subsections describe the Conditional Probability Tables (CPTs) in the form of equations for each model node. The logic behind the majority of the node equations is explained in \cite{rothmund_intention_2022}; therefore, only the new or changed node equations are explained in detail here. 
However, for completeness and implementation convenience, every node equation is at least stated.

\subsubsection{$SD_i[t]$ - Safe distance}
The parameter $SD_i[t]$ is evaluated based on the distance between the reference ship and ship $i$ at CPA and how far the reference ship crosses in front of ship $i$, assuming both ships maintain their current course and speed.
\begin{equation} \label{eq:sd_i}
    SD_i[t] = (\mathcal{M}_{DCPA_i}[t]>I_{SD}) \wedge (\mathcal{M}_{DF_i}[t]>I_{SDF})
\end{equation}

\subsubsection{$P_i[t]$ - Safely passed}
If the reference ship has passed ship $i$ $\mathcal{M}_{P_i}$ and is at a safe distance ($SD_i$), then it is considered that the reference ship has passed ship $i$ safely.
\begin{equation} \label{eq:p_i}
    P_i[t] = (\mathcal{M}_{P_i}[t]=true) \wedge SD_i[t]
\end{equation}

\subsubsection{$C\_OTen_i[t]$, $C\_OTing_i[t]$ - Correct overtaking evasive maneuver}
It is only dependent on the fact that the ships are crossing at a safe distance ($SD_i$).
\begin{equation} \label{eq:c_oten_i}
    C\_OTen_i[t] = SD_i[t]
\end{equation}
\begin{equation} \label{eq:c_oting_i}
    C\_OTing_i[t] = SD_i[t]
\end{equation}

\subsubsection{$C\_HO_i[t]$ - Correct head-on evasive maneuver}
The reference ship is performing a correct evasive maneuver if the distance to the midpoint between the ships ($\mathcal{M}_{DM_i}$) is sufficient and the ships pass the midpoint ($\mathcal{M}_{MPS_i}$) on their port side.
\begin{equation} \label{eq:c_ho_i}
    C\_HO_i[t] = (\mathcal{M}_{DM_i}[t]>I_{SDM}) \wedge (\mathcal{M}_{MPS_i}[t]==port)
\end{equation}

\subsubsection{$C\_CR\_SS_i[t]$ - Correct crossing starboard-side evasive maneuver}
The reference ship is performing a correct evasive maneuver if the reference ship is passing ship $i$ on its port side ($\mathcal{M}_{PS_i}$) at a safe distance ($SD_i$).
\begin{equation} \label{eq:c_cr_ss_i}
    C\_CR\_SS_i[t] = SD_i[t] \wedge (\mathcal{M}_{PS_i}[t]==port)
\end{equation}

\subsubsection{$C\_CR\_PS_i[t]$ - Correct crossing port-side evasive maneuver}
The reference ship is performing a correct evasive maneuver if the reference ship is passing ship $i$ at a safe distance ($SD_i$), and avoid changing its course ($CIC_i$) towards port.
\begin{equation} \label{eq:c_cr_ps_i}
    C\_CR\_PS_i[t] = SD_i[t] \wedge (\mathcal{M}_{CIC_i}[t]\neq port)
\end{equation}

\subsubsection{$NAV_M[t]$ - Navigational maneuver}
The reference ship is heading towards the next waypoint if the relative bearing ($\mathcal{M}_{WPRB}$) is decreasing, or if the waypoint is ahead of the vessel's course ($\mathcal{M}_{WPAH}$), while the distance to the next waypoint ($\mathcal{M}_{WPRD}$) is decreasing. 
\begin{dmath}[compact] \label{eq:nav_m}
    NAV\_M[t] = (\mathcal{M}_{WPRD}[t]==decreasing) \\ \wedge (\mathcal{M}_{WPRB}[t]==decreasing \\ \vee \mathcal{M}_{WPAH}[t])
\end{dmath}

\begin{figure}[htb]
    \centering
    \includegraphics[width=0.3\textwidth]{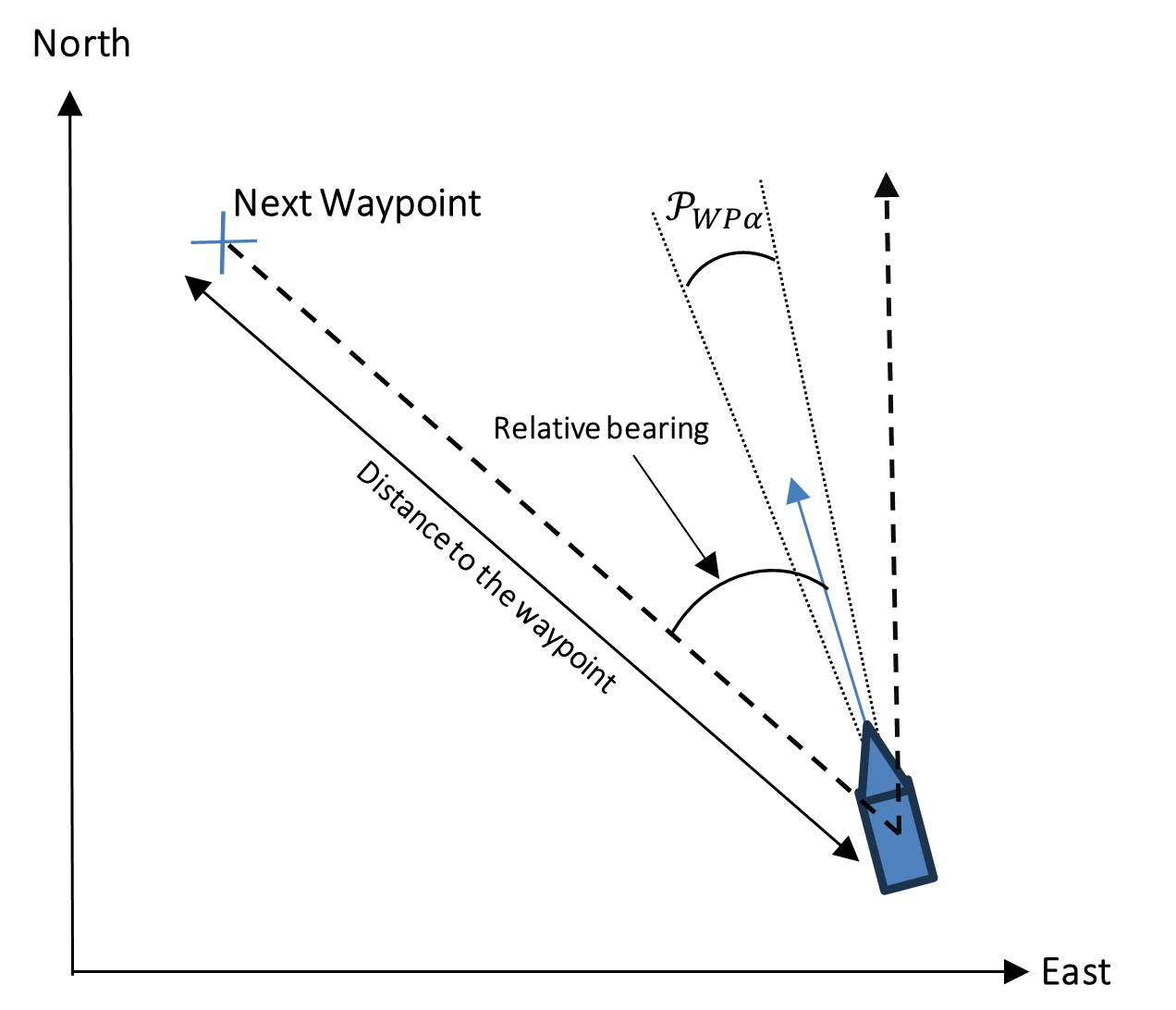}
    \caption{Graphical representation of the parameters relative bearing (used to determine $\mathcal{M}_{WPRB}$), distance to waypoint (used to determine $\mathcal{M}_{WPRD}$) and $\mathcal{P}_{WP\alpha}$.}
    \label{fig:wpm}
\end{figure}

The $\mathcal{M}_{WPRB}$ measurement considers relative bearing difference over the last $\mathcal{P}_{WPT}$ seconds while $\mathcal{M}_{WPRD}$ considers the distance change over the same time period. With the measurement $\mathcal{M}_{WPAH}$, it is checked whether the next waypoint is within $\mathcal{P}_{WP\alpha}$ degree range from its heading. These parameters are illustrated in \cref{fig:wpm}.

\subsubsection{$C\_NAV\_M_i[t]$ - Correct navigational maneuver}
The reference ship is performing a correct navigational maneuver when the navigational maneuver ($NAV\_M$) result in passing ship $i$ at a safe distance ($SD_i$) or when the ship has already passed ship $i$ ($\mathcal{M}_{P_i}$).
\begin{dmath}[compact] \label{eq:c_nav_m_i}
    C\_NAV\_M_i[t] = NAV\_M[t] \wedge \\
    \Big((SD_i[t]\wedge I_{CS_i} \neq HO)  \vee \mathcal{M}_{P_i}[t] \\
    \vee (\mathcal{M}_{DM_i}[t]>I_{SDM} \wedge I_{CS_i} = HO) \Big)
\end{dmath}

\subsubsection{$R_i[t]$ - Role}
The reference ship's role towards ship $i$ is defined as,
\begin{dmath}[compact] \label{eq:r_i}
    R_i[t]= 
\begin{cases}
    GW,& \\ \text{if} (I_{P_i}==lower) \vee \biggl((I_{P_i}==similar)\\ \wedge \Bigl((I_{CS_i}==HO) \vee (I_{CS_i}==CR\_SS)\\ \vee (I_{CS_i}==OT\_ing)\Bigr) \biggr)\\
    SO,& \text{otherwise}
\end{cases}
\end{dmath}

\subsubsection{$GS_i[t]$ - Good seamanship}
If the reference ship changed its course to port then it has to pass with ship i on its starboard side, and vice versa, in addition to not being allowed to perform both a starboard action ($SA[t]$) and a port action ($PA[t]$).
\begin{equation} \label{eq:gs_i}
    GS_i[t] = \neg (SA[t] \wedge PA[t]) \wedge (\mathcal{M}_{CIC}[t]\neq \mathcal{M}_{PS_i}[t])
\end{equation}

\subsubsection{$CEM_i[t]$ - Correct evasive maneuver}
For an evasive maneuver to be correct, the reference ship must follow good seamanship ($GS_i$) if intended ($I_{GS}$) and comply with COLREGs if it intends to be COLREGs-compliant ($I_{CC}$).
\begin{dmath}[compact] \label{eq:cem_i}
    CEM_i[t] = (\neg I_{GS}\vee GS_i[t])\wedge 
    \\  \biggl( \neg I_c \vee 
                        \Bigl( 
                                (I_{CS_i}==OT\_ing \wedge C\_OTing_i[t]) 
                                \\ \vee (I_{CS_i}==OT\_en \wedge C\_OTen_i[t]) 
                                \\ \vee (I_{CS_i}==HO \wedge C\_HO_i[t]) 
                                \\ \vee (I_{CS_i}==CR\_SS \wedge C\_CR\_SS_i[t]) 
                                \\ \vee (I_{CS_i}==CR\_PS \wedge C\_CR\_PS_i[t]) 
                        \Bigr) 
        \biggr)
\end{dmath}

\subsubsection{$SOC_i[t]$ - Stands on correctly}
The reference ship stands on correctly toward ship $i$ if it maintains course ($\mathcal{M}{CIC}$) and speed ($\mathcal{M}{CIS}$) or performs a correct evasive maneuver ($CEM_j$) toward another ship ($j$) to which it has a give-way role ($R_j$) and hasn't yet passed ($P_j$).
\begin{dmath}[compact] \label{eq:soc_i}
    SOC_i[t] = (\mathcal{M}_{CIC}[t]==straight \wedge \mathcal{M}_{CIS}[t]==none) \\ \vee_{j=1}^{n} (R_j[t]==GW \wedge CEM_j[t] \wedge \neg P_j[t])
\end{dmath}

\subsubsection{$SDG[t]$ - Safe distance to the ground}
\begin{figure}[htb]
    \centering
    \includegraphics[width=0.15\textwidth]{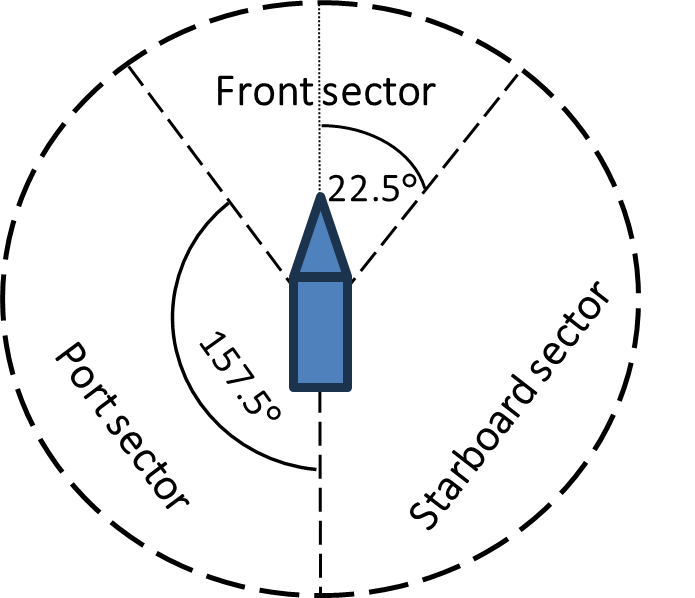}
    \caption{The three conceptual zones surrounding a reference vessel when considering grounding hazards.}
    \label{fig:gdm}
\end{figure}

To evaluate $SDG$, the area around the vessel is divided into three zones, namely, starboard, port and front as shown in \cref{fig:gdm}. 
First, safe distance to the ground from side ($SDG\_S$) is considered by calculating the current distance from reference ship to the closest point on ground in both starboard ($\mathcal{M}_{DGSB}$) and port ($\mathcal{M}_{DGPS}$) sectors separately as shown in Eq. \eqref{eq:sdg_s}. When the vessel course ($\mathcal{M}_{CIC}$) is changing to the startboard side, $\mathcal{M}_{DGSB}$ is considered while $\mathcal{M}_{DGPS}$ when changing to port. When the reference ship is keeping a straight course, $SDG\_S$ is considered 'true'.
\begin{dmath}[compact] \label{eq:sdg_s}
    SDG\_S[t] =  
   (\mathcal{M}_{DGSB}[t]>I_{SDGS} \\ \wedge \mathcal{M}_{CIC}[t]==starboard) \ \ \ \ \ \\ \vee (\mathcal{M}_{DGPS}[t]>I_{SDGS} \wedge \mathcal{M}_{CIC}[t]==port)\\ \vee (\mathcal{M}_{CIC}[t]==straight)
\end{dmath}
Next, safe distance to the ground from front ($SDG\_F$) is considered by calculating the current distance from reference ship to the closest point on the ground in the front sector as,
\begin{equation} \label{eq:sdg_f}
    SDG\_F[t] = \mathcal{M}_{DGF}[t]>I_{SDGF} \vee \mathcal{M}_{CIC}[t] \neq straight
\end{equation}
Finally, using $SDG\_S$ and $SDG\_F$, the definition for node $SDG$ can be derived as,
\begin{equation} \label{eq:sdg}
    SDG[t] = SDG\_S[t] \wedge SDG\_F[t]
\end{equation}

\subsubsection{$GWC_i[t]$ - Gives way correctly}
The reference ship gives way correctly to ship $i$ if it performs an evasive maneuver ($CEM_i$) in ample time ($I_{AT}$). It may stand on correctly ($SOC_i$) before that. Additionally, it gives way correctly if it is currently changing course ($\mathcal{M}{CCC}$) \cite{tengesdal_obstacle_2024}, where $\mathcal{M}{CCC}$ is true if the course has changed by more than $\mathcal{P}{CIC}$ degrees in the last $\mathcal{P}{CCT}$ seconds.
\begin{dmath}[compact] \label{eq:gwc_i}
    GWC_i[t] = CEM_i[t] \vee (\mathcal{M}_{CCC}[t]=true)\\ \vee \Bigl( (\mathcal{M}_{TCPA_i}[t]>I_{AT}) \wedge SOC_i[t] \Bigr) 
\end{dmath}

\subsubsection{$C\_COLAV\_M_i[t]$ - Correct collision avoidance maneuver}
Reference ship is performing a correct collision avoidance maneuver if it stand-on correct ($SOC_i$) when its role ($R_i$) is to stand-on or give way correct ($GWC_i$) when its role is to give way before passing ship $i$.
\begin{dmath}[compact] \label{eq:c_colav_m_i}
    C\_COLAV\_M_i[t] = \Bigl(\neg P_i[t] \wedge (R_i[t]==SO) \wedge SOC_i[t]\Bigr) \\ \vee \\ \Bigl(\neg P_i[t] \wedge (R_i[t]==GW) \wedge GWC_i[t]\Bigr)
\end{dmath}

\subsubsection{$C_i[t]$ - Compatible towards ship $i$}
An observation is compatible with the intention states of the reference ship towards ship $i$ if both of the following conditions are true:
\begin{itemize}
    \item The reference ship is either performing a correct collision avoidance maneuver towards ship $i$ ($C\_COLAV\_M_i$) or correct navigational maneuver ($C\_NAV\_M_i$) is performed.
    \item The vessel maintain a safe distance from the ground ($SDG$) or when it intends to sail towards the ground ($I_G$). 
\end{itemize}
Hence the node definition can be set as,
\begin{dmath}[compact] \label{eq:c_i}
    C_i[t] = (C\_COLAV\_M_i[t] \vee C\_NAV\_M_i[t]) \wedge (SDG_i[t] \vee I_G)
\end{dmath}

\subsubsection{$C[t]$ - Compatible to all} 
An observation is compatible with the reference ship's intention if it is compatible with all ships in the area ($C_i$) or if the ship intends to act in an unmodeled manner ($I_U$).
\begin{equation} \label{eq:c}
    C[t] = (\wedge_{i=1}^{n} C_i[t]) \vee I_U
\end{equation}

\subsection{Measurements for candidate trajectory evaluation} \label{subsubsec:measurements_candidate_traj}

% Some measurement nodes require minor alterations to evaluate the state of measurements based on candidate trajectories. The time until CPA ($\mathcal{M}_{TCPA_i}$) is not relevant for candidate trajectories and instead this measurement is set to minimum acceptable time ($\mathcal{P}_{AT_{min}}$) \cite{rothmund_intention_2022}. The change in course ($\mathcal{M}_{CIC}$) and change in speed ($\mathcal{M}_{CIS}$) parameters are evaluated comparing a point a bit into the trajectory with the initial point at the start of the situation \cite{rothmund_intention_2022}. The measurement accounting to whether the reference ship is currently changing course ($\mathcal{M}_{CCC}$) is set to $false$ during trajectory evaluation \cite{tengesdal_obstacle_2024}.

Some measurement nodes require minor adjustments to calculate the state of measurements when evaluating candidate trajectories. For example, parameters such as the time until CPA ($\mathcal{M}_{TCPA_i}$), change in course ($\mathcal{M}_{CIC}$), change in speed ($\mathcal{M}_{CIS}$) (explained in \cite{rothmund_intention_2022}), and whether the reference ship is changing course ($\mathcal{M}_{CCC}$) (explained in \cite{tengesdal_obstacle_2024}) require these adjustments. 

The measurements relative bearing difference ($\mathcal{M}_{WPRB}$) and distance difference ($\mathcal{M}_{WPRD}$) is evaluated considering the starting point of the trajectory and a point a bit into the trajectory. With measurement $\mathcal{M}_{WPAH}$, it is checked whether the vessel at a point further into the trajectory has the next waypoint within $\mathcal{P}_{WP\alpha}$ degree range from its heading. Similarly, $M_{DGSB}$, $M_{DGPS}$ and $M_{DGF}$ are also evaluated considering the reference vessel at a point further into the trajectory.

\section{Parameter Extraction from AIS Data} \label{sec:parameter_extraction}
% Using the filtered data, preliminary studies have been conducted by \cite{rothmund_risk_2023,veglo_ais_2022,haugen_validation_2023} to extract parameter values of the previous intention models.
It is essential to define appropriate values for the tunable parameters of the DBN. The prior probability distributions of the intention nodes stand out, as they have a direct impact on the probability predictions of the intention model. These can be determined through various methods, such as expert workshops, observing ship captains, or analyzing AIS data \cite{rothmund_intention_2022,rothmund_validation_2023}. For this research, we decided to extract these distributions from historical AIS data. 

The AIS dataset was provided by the Norwegian Coastal Administration and was collected from 2018 to 2021. An analysis to filter, isolate, and categorize vessel encounters in the dataset was carried out by \cite{hagen_exploration_2024}. From the filtered data, 1,377 encounters were manually inspected and verified, and these are used in this paper for parameter extraction of the proposed intention model. This dataset consists of 770 head-on, 258 overtaking, and 349 crossing encounters.

Next, to extract parameters related to grounding hazards in the model, grounding hazard information surrounding each encounter was obtained from \textit{The Fjord Catalog} published by \cite{the_norwegian_environment_agency_miljodirektoratet_2025}. The shapefile data included in this dataset was used to parameterize grounding hazards as two-dimensional polygons. Then, using polygon manipulation libraries, the required parameter values were calculated.

The extraction process of the prior probability distribution for each intention node is explained in detail in the following subsections. The distributions extracted using these processes from the said dataset are listed in \ref{sec:app1}.

\subsection{$I_{SDF}$ Probability Distribution} \label{subsec:Isf}
The node $I_{SDF}$ represent how far in front of a ship the reference ship considers a crossing as safe. In the DBN, this parameter directly influence $C\_CR\_SS_i$ and $C\_CR\_PS_i$ probabilities. In other words, this parameter corresponds mainly to the crossing situations defined in COLREGS Rule 15 \cite{imo_convention_1972}. Hence, to extract the probability distribution, only situations classified as crossing in the AIS dataset have been used. The safest distance in front at crossing does not necessarily occur at the CPA. Therefore, we propose calculating the distance between the vessels at every instance of the trajectory in an encounter when the obstacle vessel is in front of the reference vessel and taking the minimum as the safe distance for that encounter. To validate whether the obstacle vessel is in front, we check if it lies within the limits of a small sector in the front portion of the reference vessel's ship domain.

Algorithm \ref{alg:find_isdf_vals} is used for this task. For every encounter in the dataset, the reference ship's trajectory $refship\_traj$, and the obstacle ship's trajectory $obstship\_traj$, are first extracted. A variable, $min\_dist$, is initialized to track the minimum distance identified during the search process. At each time step of the two trajectories, the heading vector of the reference ship $head\_vec$, and the relative position vector between the two ships $relpos\_vec$, are computed. The normalized dot product of these vectors is then evaluated. If the value is greater than $\cos(\pi/8)$, it means that the obstacle ship is located within a $45^\circ$ sector ($\pm22.5^\circ$) around the reference ship's heading direction. If this condition is met, the Euclidean distance between the two ships at that time step is calculated. If the calculated distance is shorter than the current $min\_dist$, the $min\_dist$ is updated accordingly. Once the $min\_dist$ for a specific encounter is identified, it is appended to the output variable $Isdf\_vals$.
\begin{algorithm}[htb]
\caption{\textit{find\_Isdf\_vals()} \\ Calculate the shortest distance at which the obstacle ship is located directly in front of the reference ship for each encounter.}\label{alg:find_isdf_vals}
\KwData{AIS data for each encounter $encounter\_data$}
\KwResult{$Isdf\_vals$}
\ForEach{$encounter$ in $encounter\_data$}{
    $refship\_traj \gets encounter$\;
    $obstship\_traj \gets encounter$\;
    $min\_dist \gets \infty$\;
    \ForEach{$refship\_state$ and $obstship\_state$ in $refship\_traj$ and $obstship\_traj$}{
        $x\_ref, y\_ref, chi\_ref \gets refship\_state$\;
        $x\_obst, y\_obst \gets obstship\_state$\;

        $dx = x\_obst - x\_ref$\;
        $dy = y\_obst - y\_ref$\;

        $head\_vec = [cos(chi\_ref), sin(chi\_ref)]$\;
        $relpos\_vec = [dx, dy]$\;
        \If{$\dfrac{\text{dot}(head\_vec,relpos\_vec)}{\left \lVert relpos\_vec \right \rVert \cdot \left \lVert head\_vec \right \rVert} > \cos(\pi/8)$}{
            $dist = \sqrt{dx^2+dy^2}$\;
            \If{$dist < min\_dist$}{
                $min\_dist=dist$\;
            }
        }
    }
    \If{$min\_dist \neq \infty$}{
        $Isdf\_vals \gets min\_dist$\;
    }
}
\end{algorithm}

\subsection{$I_{SD}$,$I_{SDM}$ and $I_{AT}$ Probability Distributions} \label{subsec:Isd_Isdm_Iat}
The nodes $I_{SD}$ and $I_{SDM}$ represent the safe distance between the vessels and the safe distance to the current midpoint at CPA, respectively, as considered by the reference ship. $I_{AT}$ represents the time until CPA that the reference ship considers ample. To calculate the prior probability distributions for these intention nodes, Algorithm \ref{alg:findcpa} is employed. The algorithm takes encounter data as input, with the goal of determining the Distance at Closest Point of Approach (DCPA) and Time to Closest Point of Approach (TCPA) for each encounter in the dataset. Within a loop, each encounter in the dataset is extracted and individually assessed. The considered encounters differ depending on the three intention nodes. For $I_{SD}$ only overtaking categorized encounters are used while only encounters categorized as head-on are used for $I_{SDM}$. Encounters of all COLREGS types are considered for $I_{AT}$.
\begin{algorithm}[htbp]
\caption{\textit{findCPA()} \\ Calculate the DCPA and TCPA for each encounter.}\label{alg:findcpa}
\KwData{AIS data for each encounter $encounter\_data$}
\KwResult{$dcpa\_vals$,$tcpa\_vals$}
\ForEach{$encounter$ in $encounter\_data$}{
    $refship\_traj \gets encounter$\;
    $obstship\_traj \gets encounter$\;
    $min\_dist \gets \infty$\;
    $dcpa \gets \infty$\;
    $tcpa \gets \infty$\;
    $t_0^{ref}, x_0^{ref}, y_0^{ref}, \chi_0^{ref} \gets refship\_traj$\;
    \ForEach{$refship\_state$ and $obstship\_state$ in $refship\_traj$ and $obstship\_traj$}{
        $dist \gets $ current euclidean distance between reference ship and obstacle ship\;
        $t_{curr}^{ref}, x_{curr}^{ref}, y_{curr}^{ref}, \chi_{curr}^{ref} \gets refship\_state$\;
        \If{$dist < min\_dist$}{
            $min\_dist = dist$\;
            $ref\_next\_state \gets $ reference ship state at the next time step\;
            $obst\_next\_state \gets $ obstacle ship state at the next time step\;
            $t^{opt},d^{opt} \gets \text{Algorithm \ref{alg:calccpaline}}$\;
            $t^{opt},d^{opt} \gets \begin{aligned}
                                        calcCPALine(  
                                                &refship\_state,\\
                                                &ref\_next\_state,\\
                                                &obstship\_state,\\
                                                &obst\_next\_state)
                                    \end{aligned}\hspace{0.5cm}$\;
            \eIf{$d < dist$}{
                $dcpa \gets d^{opt}$\;
                $tcpa = t_{curr}^{ref} - t_0^{ref} + t^{opt}$\;
            }{
                $dcpa \gets min\_dist$\;
                $tcpa = t_{curr}^{ref} - t_0^{ref}$\;
            }
        }
    }
    \If{$dcpa \neq \infty$}{
        $dcpa\_vals \gets dcpa$\;
    }
    \If{$tcpa \neq \infty$}{
        $tcpa\_vals \gets tcpa$\;
    }
    
}
\end{algorithm}

From each encounter, similar to Algorithm \ref{alg:find_isdf_vals}, the trajectories of the reference ship and the obstacle ship are extracted as $refship\_traj$ and $obstship\_traj$. Next, the variables $min\_dist$, $dcpa$, and $tcpa$ are initialized to infinity. These variables are intended to temporarily store the current minimum distance, DCPA, and TCPA values, respectively. Additionally, the initial state of $refship\_traj$ is extracted as $t_0^{ref}, x_0^{ref}, y_0^{ref}, \chi_0^{ref}$ for later use. Here, $t$, $x$, $y$, and $\chi$ represent the timestamp in UNIX format, the easting and northing coordinates in the Universal Transverse Mercator (UTM) coordinate system, and the course angle of the vessel, respectively. Then, within a nested loop, the states of both vessels at each time step of the two trajectories are accessed using the variables $refship\_state$ and $obstship\_state$.

At each iteration of the nested loop, the current state of $refship\_traj$ is extracted as $t_{curr}^{ref}, x_{curr}^{ref}, y_{curr}^{ref}, \chi_{curr}^{ref}$. Then, the Euclidean distance between the two vessels is calculated and assigned to $dist$. If $dist < min\_dist$, the $min\_dist$ variable is updated with the current distance. The goal here is to identify the vessel states at the closest points of the trajectories. However, since the trajectories are discretized with a fixed sampling time, there could be a point between two consecutive vessel positions samples where the vessels are closer to each other than at the closest discretized points in the dataset. Hence, each time step is further evaluated using Algorithm \ref{alg:calccpaline}.
\begin{algorithm}[htb]
\caption{\textit{calcCPALine()} \\ Calculate the point where the vessels are the closest when sailing between two consecutive AIS data positions.}\label{alg:calccpaline}
\KwData{$v_1\_curr\_state, v_1\_next\_state,$\ $v_2\_curr\_state, v_2\_next\_state$}
\KwResult{Time until the point $t^{opt}$ and the distance at the point $d^{opt}$}
$x_1^{curr},y_1^{curr},u_1^{curr},t_1^{curr} \gets v_1\_curr\_state$\;
$x_1^{next},y_1^{next},u_1^{next},t_1^{next} \gets v_1\_next\_state$\;
$x_2^{curr},y_2^{curr},u_2^{curr},t_2^{curr} \gets v_2\_curr\_state$\;
$x_2^{next},y_2^{next},u_2^{next},t_2^{next} \gets v_2\_next\_state$\;
$\chi_1=\operatorname{arctan}\bigg(\dfrac{y_1^{next}-y_1^{curr}}{x_1^{next}-x_1^{curr}}\bigg)$\;
$\chi_2=\operatorname{arctan}\bigg(\dfrac{y_2^{next}-y_2^{curr}}{x_2^{next}-x_2^{curr}}\bigg)$\;
$pos_1=
\begin{bmatrix}
x_1^{curr}+(u_1^{curr} * \cos(\chi_1) * t)\\
y_1^{curr}+(u_1^{curr} * \sin(\chi_1) * t)
\end{bmatrix}$\;
$pos_2=
\begin{bmatrix}
x_2^{curr}+(u_2^{curr} * \cos(\chi_2) * t)\\
y_2^{curr}+(u_2^{curr} * \sin(\chi_2) * t)
\end{bmatrix}$\;
$dist = \lVert pos_2 - pos_1 \rVert$\;
$t^{opt} \gets \text{Solve for } t \text{ where } dist \text{ is at its minimum}$\;
$pos_1^{opt} \gets \text{Solve } pos_1 \text{ equation using } t=t^{opt}$\;
$pos_2^{opt} \gets \text{Solve } pos_2 \text{ equation using } t=t^{opt}$\;
$d^{opt} \gets \text{Solve } dist \text{ equation using } pos_1^{opt} \text{ and } pos_2^{opt}$\;
\end{algorithm}

In Algorithm \ref{alg:calccpaline}, each vessel's current and next state is taken as input to calculate the point where the vessels are closest between two consecutive AIS data points. First, an equation is formulated for the distance between the two vessels when sailing between two consecutive AIS points at a random point in time $t$. Then, a solver is used to find $t = t^{opt}$, which gives the minimum Euclidean distance between the vessels. Finally, using $t^{opt}$, the distance between the vessels at the closest point, $d^{opt}$, can be calculated.

The values $t^{opt}$ and $d^{opt}$ are then used in Algorithm \ref{alg:findcpa} to calculate $dcpa$ and $tcpa$ for each encounter. These values are subsequently appended to the output variables $dcpa\_vals$ and $tcpa\_vals$, which store the respective values for each encounter. The $dcpa\_vals$ for overtaking encounters are used to calculate the probability distribution of $I_{SD}$, while $dcpa\_vals / 2$ for head-on encounters are used for $I_{SDM}$. Regardless of the COLREGS situation, the $tcpa\_vals$ for all encounters are used to calculate the probability distribution of $I_{AT}$.

\subsection{$I_{SDGS}$ and $I_{SDGF}$ Probability Distributions} \label{subsec:Isdgs_Isdgf}
Nodes $I_{SDGS}$ and $I_{SDGF}$ represent what the reference ship considers a safe distance from either side of the vessel and from the front to the ground, respectively. To find the probability distributions for these nodes, Algorithm \ref{alg:find_dist2grd_cpa} is used. The inputs to the algorithm are the AIS data for the encounters, denoted as $encounter\_data$, and the map information represented as polygons, denoted as $map\_poly$. A Region of Interest (ROI) of 10 km around the encounter CPA is then extracted from $map\_poly$ for consideration. Next, the closest distances to the ground in the starboard ($sb\_dist$), port ($ps\_dist$), and front ($fr\_dist$) sectors of the reference ship's domain are calculated using Algorithm \ref{alg:calc_dist2grd_in_sect}. These sectors align with the segmentation of the ship domain explained in \cref{fig:gdm}.
\begin{algorithm}[htbp]
\caption{\textit{find\_dist2grd\_cpa()} \\ Calculate the closest distance to ground at CPA of the encounter.}\label{alg:find_dist2grd_cpa}
\KwData{AIS data for each encounter $encounter\_data$, $map\_poly$}
\KwResult{$sdgs\_vals$, $sdgf\_vals$}
\ForEach{$encounter$ in $encounter\_data$}{
    $refship\_traj \gets encounter$\;
    $x_{cpa}^{ref}, y_{cpa}^{ref}, \chi_{cpa}^{ref} \gets refship\_state$\;
    $roi\_len = 10km$\;
    $roi^{min}_x = x_{cpa}^{ref} - roi\_len$\;
    $roi^{max}_x = x_{cpa}^{ref} + roi\_len$\;
    $roi^{min}_y = y_{cpa}^{ref} - roi\_len$\;
    $roi^{max}_y = y_{cpa}^{ref} + roi\_len$\;
    $roi\_poly \gets \text{polygon using } roi^{min}_x, roi^{min}_y, roi^{max}_x, roi^{max}_y$\;
    $map\_poly\_roi \gets map\_poly \cap roi\_poly$\;
    $sb\_dist = \begin{aligned}
                    calc\_dist2grd\_in\_sect(  
                        &x_{cpa}^{ref}, y_{cpa}^{ref}, \chi_{cpa}^{ref},\\
                        &map\_poly\_roi,\\
                        &\chi_{cpa}^{ref} - \pi, \chi_{cpa}^{ref} - \dfrac{\pi}{8})
                \end{aligned}\hspace{0.5cm}$\;
    $ps\_dist = \begin{aligned}
                    calc\_dist2grd\_in\_sect(  
                        &x_{cpa}^{ref}, y_{cpa}^{ref}, \chi_{cpa}^{ref},\\
                        &map\_poly\_roi,\\
                        &\chi_{cpa}^{ref} + \dfrac{\pi}{8}, \chi_{cpa}^{ref} + \pi)
                \end{aligned}\hspace{0.5cm}$\;
    $fr\_dist = \begin{aligned}
                    calc\_dist2grd\_in\_sect(  
                        &x_{cpa}^{ref}, y_{cpa}^{ref}, \chi_{cpa}^{ref},\\
                        &map\_poly\_roi,\\
                        &\chi_{cpa}^{ref} - \dfrac{\pi}{8}, \chi_{cpa}^{ref} + \dfrac{\pi}{8})
                \end{aligned}\hspace{0.5cm}$\;
    \If{$\min(sb\_dist, ps\_dist) <= dist\_thresh$}{
        $sdgs\_vals \gets \min(sb\_dist, ps\_dist)$\;
    }
    \If{$fr\_dist <= dist\_thresh$}{
        $sdgf\_vals \gets fr\_dist$\;
    }
}
\end{algorithm}

In Algorithm \ref{alg:calc_dist2grd_in_sect}, it is first checked whether the current vertex under consideration is in the vessel domain sector defined by the angles $sect_\alpha^{start}$ and $sect_\alpha^{end}$. If the condition is satisfied, the distance between the vessel's position, denoted as ($vessel_x$, $vessel_y$), and each vertex in the map polygons within the ROI ($map\_poly\_roi$) is calculated, and the minimum distance ($min\_dist$) is determined. Once the values of $sb\_dist$, $ps\_dist$, and $fr\_dist$ are determined using Algorithm \ref{alg:calc_dist2grd_in_sect}, they are used in Algorithm \ref{alg:find_dist2grd_cpa} to compute the minimum distance to the ground from the sides ($sdgs\_vals$) and front ($sdgf\_vals$) of the vessel. For this, only values less than a certain threshold ($dist\_thresh$) are considered to include only encounters with grounding hazards that affected the CPA of the encounter.
\begin{algorithm}[htbp]
\caption{\textit{calc\_dist2grd\_in\_sect()} \\ Calculate the closest distance to ground in sector in the vessel domain}\label{alg:calc_dist2grd_in_sect}
\KwData{AIS data for each encounter $vessel_x$, $vessel_y$, $vessel_\chi$, $map\_poly\_roi$, $sect_\alpha^{start}$, $sect_\alpha^{end}$}
\KwResult{$min\_dist$}
\ForEach{$vertex$ in $map\_poly\_roi$}{
    $vertex_x, vertex_y \gets vertex$\;
    $rel_x = vertex_x - vessel_x$\;
    $rel_y = vertex_y - vessel_y$\;
    $rel_\chi = \operatorname{arctan}\bigg(\dfrac{rel_y}{rel_x}\bigg)$\;
    $min\_dist \gets \infty$\;
    \If{$sect_\alpha^{start} \leq rel_\chi \leq sect_\alpha^{end}$}{
        $dist = \lVert vertex_{x,y} - vessel_{x,y} \rVert$\;
        $min\_dist = \min(min\_dist, dist)$\;
    }
}
\end{algorithm}

\section{Results} \label{sec:results}
The DBN model was developed using GeNie software \cite{bayesfusion_llc_genie_2021}, a graphical user interface for constructing and simulating Bayesian Networks. It was then integrated into a ship simulator via the Python wrapper of the SMILE Engine \cite{bayesfusion_llc_smile_2021}. The simulator was built in Python and executed on a workstation equipped with an Intel Core i7-10750H CPU, 32 GB of RAM, and an NVIDIA RTX 2070 SUPER Mobile GPU. 
% The proposed approach was extensively and successfully tested in various scenarios, though only a subset of the results is presented for conciseness.

\begin{figure}[htb]
    \centering
    \includegraphics[width=0.4\textwidth]{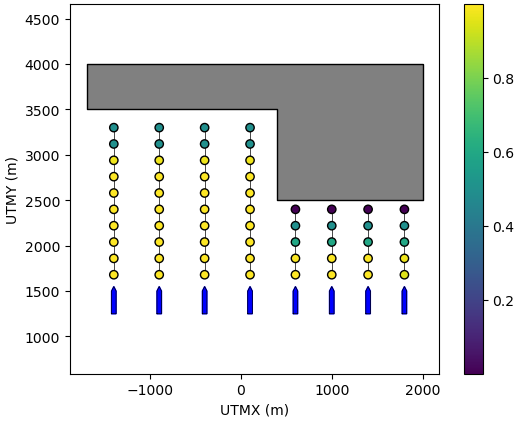}
    \caption{Multiple simulations of the reference ship approaching a grounding hazard with different sideways offsets. The figure illustrates how the $SDG\_F=true$ develop over time.}
    \label{fig:sdgf}
\end{figure}

The functionality of the newly added DBN features was tested first. \cref{fig:sdgf} illustrates how intentions develop when the reference vessel sails toward a grounding hazard with different sideways offsets over multiple simulations. The reference ship is maintaining its initial course and speed throughout each simulation. Since $M_{CIC}[t]=\text{straight}$ throughout each trajectory in the simulations, according to Eq. \eqref{eq:sdg_f}, the probability of $SDG\_F[t]=\text{true}$ gradually decreases over time as the vessel approaches the grounding hazard, as expected.

\begin{figure}[htb]
    \centering
    \includegraphics[width=0.435\textwidth]{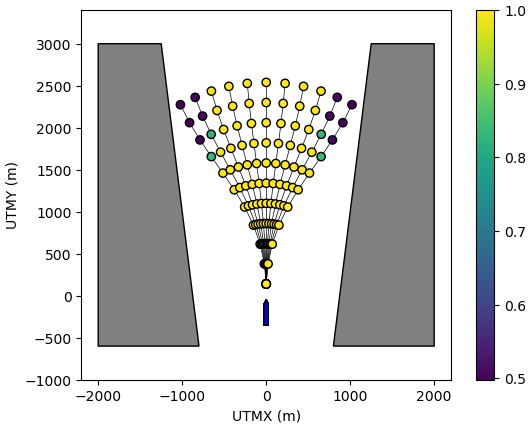}
    \caption{Multiple simulations of the reference ship approaching a grounding hazard with different course offsets. The figure illustrates how the $SDG\_S=true$ develop over time.}
    \label{fig:sdgs}
\end{figure}

Similarly, according to Eq. \eqref{eq:sdg_s}, when the vessel is turning to port ($\mathcal{M}_{CIC}[t]=port$) or starboard ($\mathcal{M}_{CIC}[t]=starboard$), and the distance to the closest point on the ground in the respective conceptual zone (as explained in \cref{fig:gdm}) decreases, the probability that $SDG\_S[t]=true$ also decreases. This is illustrated in \cref{fig:sdgs}. 

It is worth noting that in \cref{fig:sdgf,fig:sdgs}, the change in probability depends on the prior probability distribution of the $I_{SDGF}$ and $I_{SDGS}$ intention nodes, the discretization resolution of these distributions, and the input ranges of $\mathcal{M}_{DGF}$, $\mathcal{M}_{DGSB}$, and $\mathcal{M}_{DGPS}$. 

\begin{figure}[htb]
     \centering
     \begin{subfigure}[b]{\intentionImgSize}
         \centering
         \includegraphics[width=\textwidth]{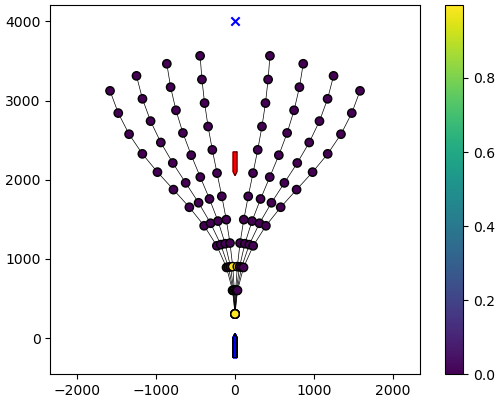}
         \caption{}
         \label{fig:cnavm-1}
     \end{subfigure}
     %\hfill
     \begin{subfigure}[b]{\intentionImgSize}
         \centering
         \includegraphics[width=\textwidth]{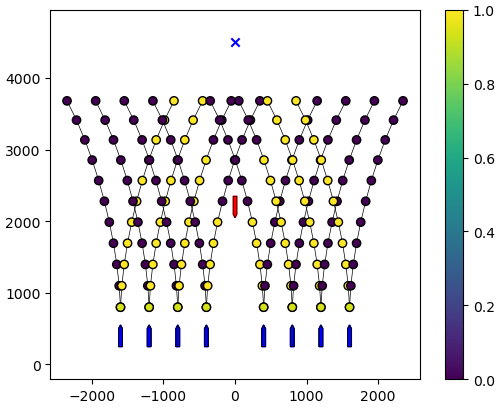}
         \caption{}
         \label{fig:cnavm-2}
     \end{subfigure}
    \caption{Multiple simulations of the reference ship (blue) approaching a target vessel (red) with different course offsets. The figure illustrates how the $C\_NAV\_M_i=true$ develop over time.}
    \label{fig:cnavm}
\end{figure}

Correct navigational maneuver $C\_NAV\_M_i[t]$ becomes $true$ if the conditions explained in Eq. \eqref{eq:c_nav_m_i} are fulfilled. To test this, as shown in \cref{fig:cnavm}, multiple simulations were carried out to observe different trajectories generated through various course offsets. In \cref{fig:cnavm}, the reference ship is depicted in blue, while the obstacle ship is in red. The reference vessel maintains its initial speed throughout its trajectory in each simulation, whereas the target vessel remains stationary. As shown in \cref{fig:cnavm-1}, any collision avoidance trajectory that moves away from the waypoint is not recognized as a navigational maneuver. \cref{fig:cnavm-2} shows how the trajectories moving towards the next waypoint are recognized as navigational maneuvers.

\begin{figure*}[htbp]
     \centering
     \begin{subfigure}[b]{\resultsImgSize}
         \centering
         \includegraphics[width=\textwidth]{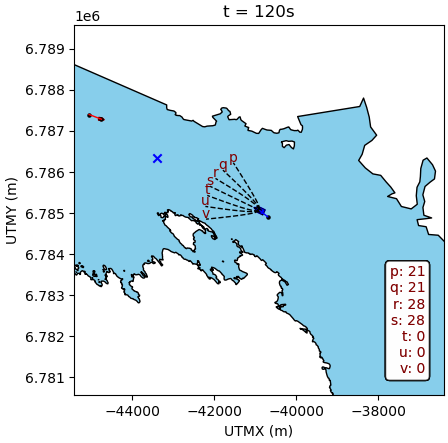}
         \caption{}
         \label{fig:29ANT-1}
     \end{subfigure}
     %\hfill
     \begin{subfigure}[b]{\resultsImgSize}
         \centering
         \includegraphics[width=\textwidth]{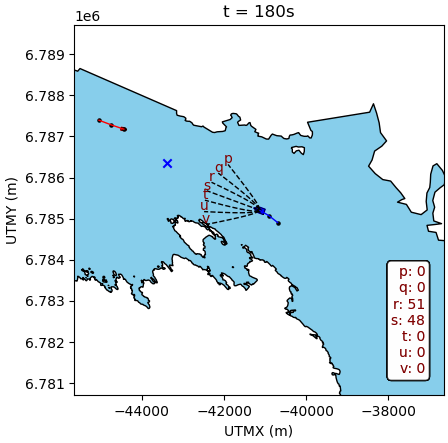}
         \caption{}
         \label{fig:29ANT-2}
     \end{subfigure}
     %\hfill
     \begin{subfigure}[b]{\resultsImgSize}
         \centering
         \includegraphics[width=\textwidth]{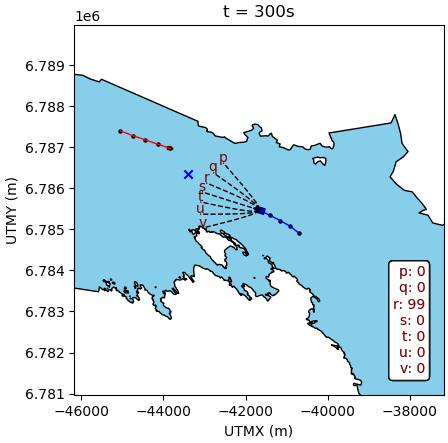}
         \caption{}
         \label{fig:29ANT-3}
     \end{subfigure}
     %\hfill
     \begin{subfigure}[b]{\resultsImgSize}
         \centering
         \includegraphics[width=\textwidth]{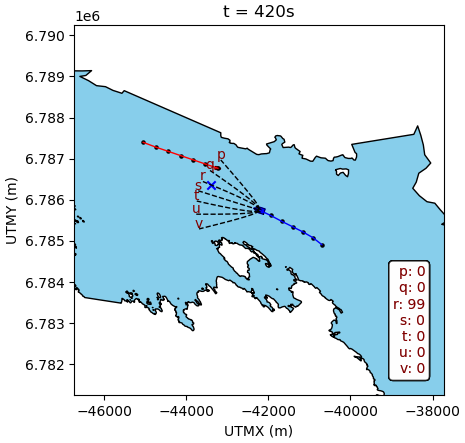}
         \caption{}
         \label{fig:29ANT-4}
     \end{subfigure}
     %\hfill
     \begin{subfigure}[b]{\inputsImgSize}
         \centering
         \includegraphics[width=\textwidth]{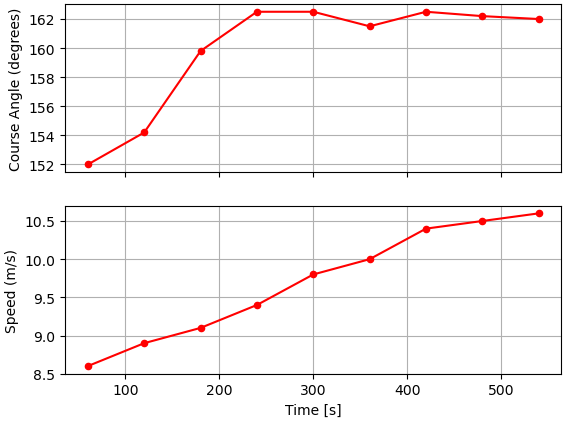}
         \caption{}
         \label{fig:29ANT-5}
     \end{subfigure}
     %\hfill
     \begin{subfigure}[b]{\probabilitiesImgSize}
         \centering
         \includegraphics[width=\textwidth]{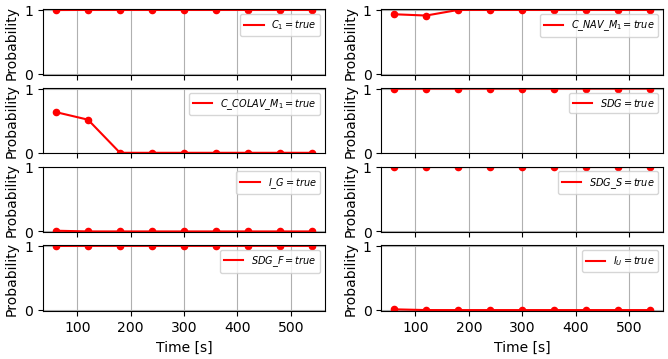}
         \caption{}
         \label{fig:29ANT-6}
     \end{subfigure}
    \caption{(a-d) Encounter (e) Course and Speed Changes (f) Probabilities of Selected Nodes from the Model. The blue ship and cross represent the reference vessel and its assumed next waypoint, while the red ship represents the obstacle vessel. $p$, $q$, $r$, $s$, and $t$ indicate the model's confidence in each candidate trajectory aligning with the intention model's expectations.}
    \label{fig:29ANT}
\end{figure*}

\begin{figure*}[htbp]
     \centering
     \begin{subfigure}[b]{\resultsImgSize}
         \centering
         \includegraphics[width=\textwidth]{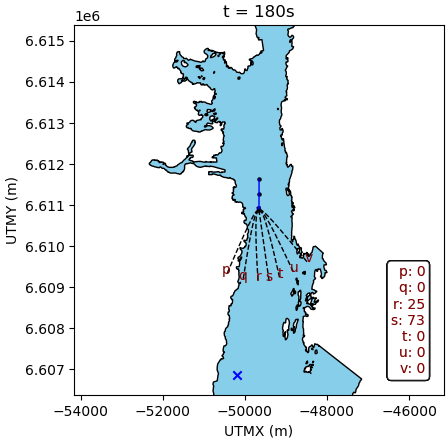}
         \caption{}
         \label{fig:SHL9A-1}
     \end{subfigure}
     %\hfill
     \begin{subfigure}[b]{\resultsImgSize}
         \centering
         \includegraphics[width=\textwidth]{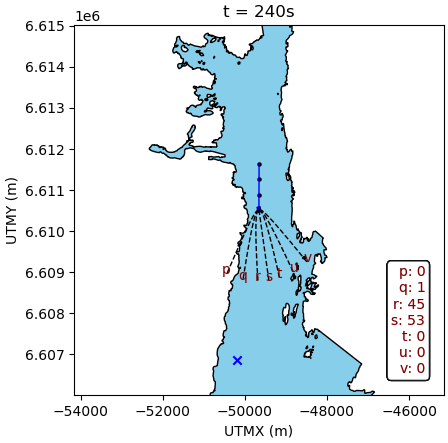}
         \caption{}
         \label{fig:SHL9A-2}
     \end{subfigure}
     %\hfill
     \begin{subfigure}[b]{\resultsImgSize}
         \centering
         \includegraphics[width=\textwidth]{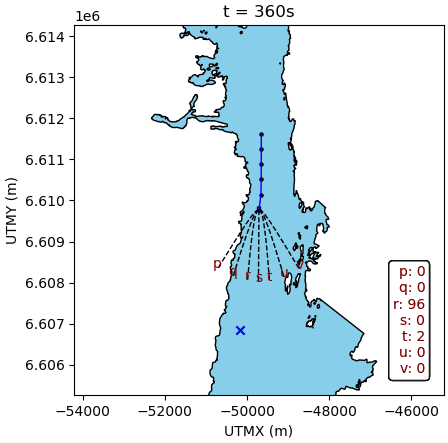}
         \caption{}
         \label{fig:SHL9A-3}
     \end{subfigure}
     %\hfill
     \begin{subfigure}[b]{\resultsImgSize}
         \centering
         \includegraphics[width=\textwidth]{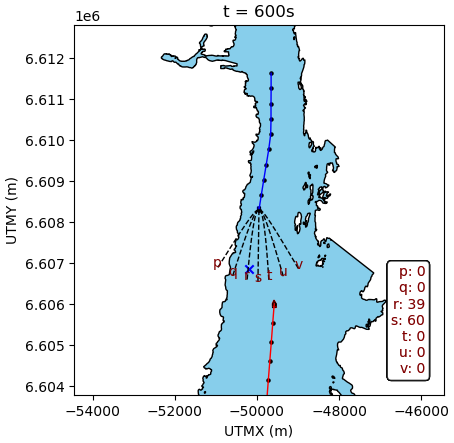}
         \caption{}
         \label{fig:SHL9A-4}
     \end{subfigure}
     %\hfill
     \begin{subfigure}[b]{\inputsImgSize}
         \centering
         \includegraphics[width=\textwidth]{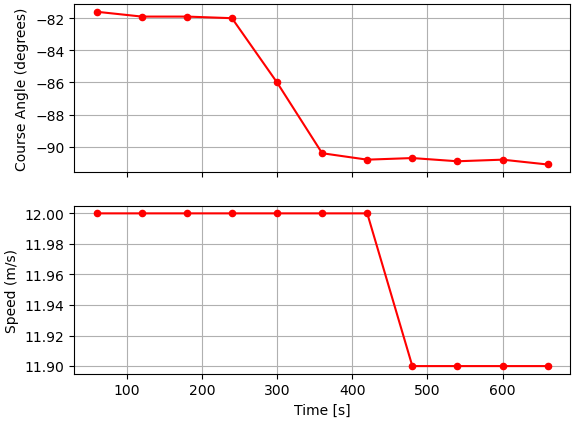}
         \caption{}
         \label{fig:SHL9A-5}
     \end{subfigure}
     %\hfill
     \begin{subfigure}[b]{\probabilitiesImgSize}
         \centering
         \includegraphics[width=\textwidth]{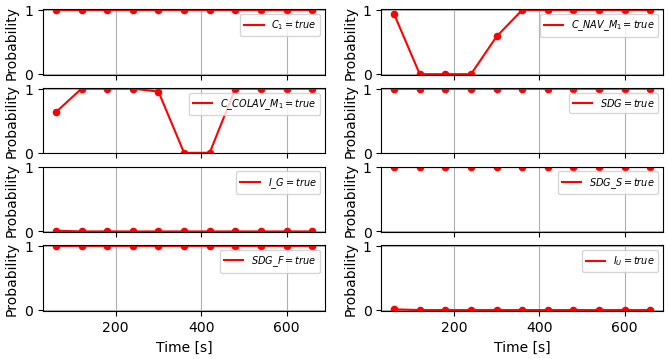}
         \caption{}
         \label{fig:SHL9A-6}
     \end{subfigure}
    \caption{(a-d) Encounter (e) Course and Speed Changes (f) Probabilities of Selected Nodes from the Model. The blue ship and cross represent the reference vessel and its assumed next waypoint, while the red ship represents the obstacle vessel. $p$, $q$, $r$, $s$, and $t$ indicate the model's confidence in each candidate trajectory aligning with the intention model's expectations.}
    \label{fig:SHL9A}
\end{figure*}

\begin{figure*}[htbp]
     \centering
     \begin{subfigure}[b]{\resultsImgSize}
         \centering
         \includegraphics[width=\textwidth]{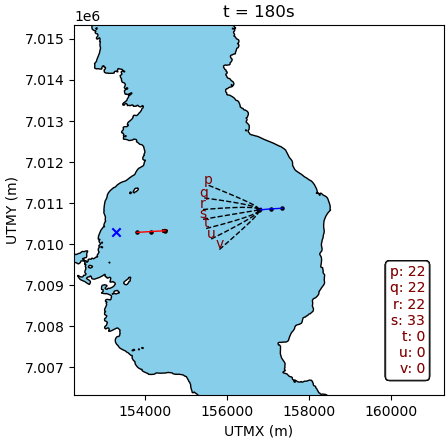}
         \caption{}
         \label{fig:G7W2E-1}
     \end{subfigure}
     %\hfill
     \begin{subfigure}[b]{\resultsImgSize}
         \centering
         \includegraphics[width=\textwidth]{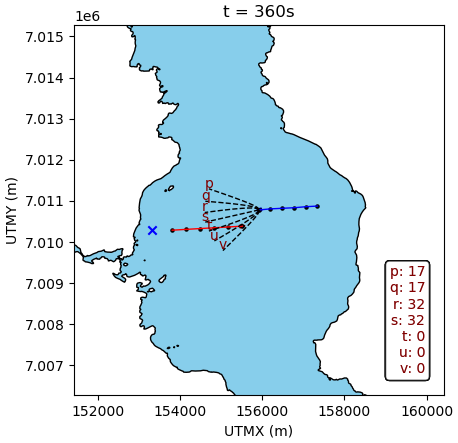}
         \caption{}
         \label{fig:G7W2E-2}
     \end{subfigure}
     %\hfill
     \begin{subfigure}[b]{\resultsImgSize}
         \centering
         \includegraphics[width=\textwidth]{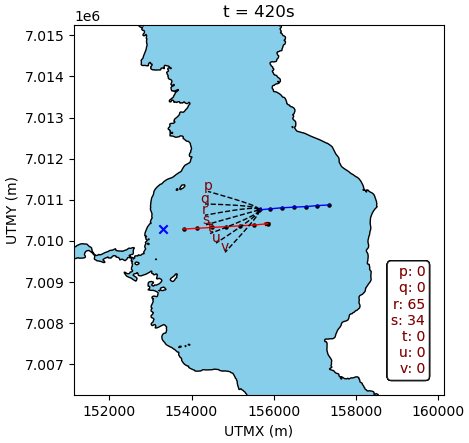}
         \caption{}
         \label{fig:G7W2E-3}
     \end{subfigure}
     %\hfill
     \begin{subfigure}[b]{\resultsImgSize}
         \centering
         \includegraphics[width=\textwidth]{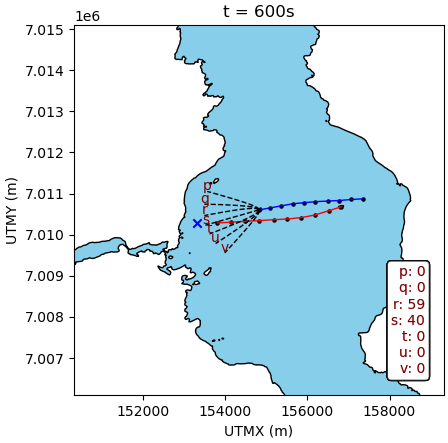}
         \caption{}
         \label{fig:G7W2E-4}
     \end{subfigure}
     %\hfill
     \begin{subfigure}[b]{\inputsImgSize}
         \centering
         \includegraphics[width=\textwidth]{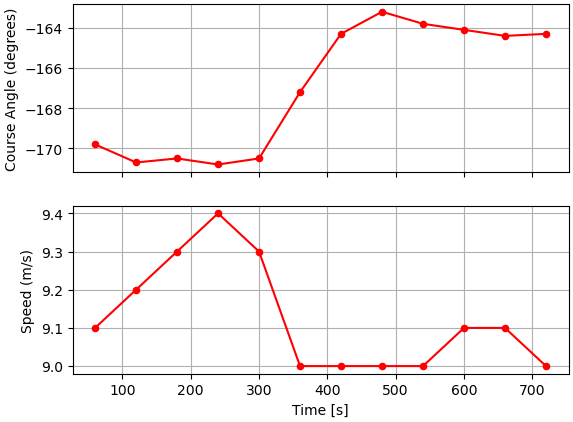}
         \caption{}
         \label{fig:G7W2E-5}
     \end{subfigure}
     %\hfill
     \begin{subfigure}[b]{\probabilitiesImgSize}
         \centering
         \includegraphics[width=\textwidth]{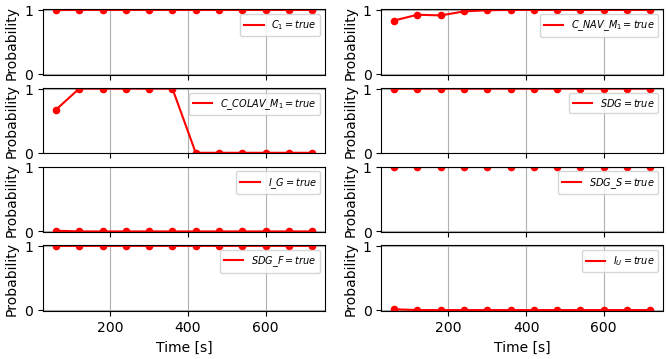}
         \caption{}
         \label{fig:G7W2E-6}
     \end{subfigure}
    \caption{(a-d) Encounter (e) Course and Speed Changes (f) Probabilities of Selected Nodes from the Model. The blue ship and cross represent the reference vessel and its assumed next waypoint, while the red ship represents the obstacle vessel. $p$, $q$, $r$, $s$, and $t$ indicate the model's confidence in each candidate trajectory aligning with the intention model's expectations.}
    \label{fig:G7W2E}
\end{figure*}

Next, to test the overall performance of the intention model, several encounters from the historical AIS dataset were selected (see \cref{fig:29ANT,fig:SHL9A,fig:G7W2E}). In each encounter, one vessel is chosen whose trajectory is to be predicted (reference vessel). The proposed intention model is implemented from the perspective of this vessel to predict its intended future motion. The trajectory data from the historical dataset is used as input to the intention model at each time step of the simulation.

In \cref{fig:29ANT,fig:SHL9A,fig:G7W2E}, the reference ship is depicted in blue, while the obstacle ship is shown in red. The blue cross represents the next waypoint of the reference ship, which is assumed for each encounter by observing the AIS data. At each time step, six candidate trajectories ($p$, $q$, $r$, $s$, and $t$) are evaluated as probable courses of action. The probability of each candidate trajectory aligning with the intention model's expectations at that moment is calculated. These six values are then normalized to represent the confidence in each trajectory as a percentage.

In the encounter shown in \cref{fig:29ANT-1,fig:29ANT-2,fig:29ANT-3,fig:29ANT-4}, a classic head-on scenario has occurred. According to COLREGS Rule 14, both vessels are required to give way so that each passes on the port side of the other. However, in this specific instance, the reference vessel is ignoring its give-way responsibility in order to reach its next waypoint. At $t=120s$, the vessels are in a state where the reference ship could either be executing a navigation maneuver or a collision avoidance maneuver. However, since the vessels are in close proximity and the next waypoint lies directly ahead, the probability of it being a collision avoidance maneuver ($C\_COLAV\_M_1$) is lower compared to that of being a navigation maneuver ($C\_NAV\_M_1$), as shown in \cref{fig:29ANT-6}. At $t=180s$, the probability of $C\_COLAV\_M_1$ drops to zero due to a drastic change in the course angle towards port, as shown in \cref{fig:29ANT-5}. This indicates that the model is confident that the reference maneuver is a navigation maneuver towards its next waypoint. This is further highlighted in \cref{fig:29ANT-3,fig:29ANT-4}, where the candidate trajectories moving towards the next waypoint have the highest confidence percentage.

In the encounter shown in \cref{fig:SHL9A-1,fig:SHL9A-2,fig:SHL9A-3,fig:SHL9A-4}, a similar encounter has occurred in a more congested environment with closer grounding hazards. However, in this instance, the starboard maneuver to navigate to the next waypoint aligns with the give-way responsibility of the reference vessel. This is explained by the probability shift in $C\_NAV\_M_1$ after $t=240s$ (see \cref{fig:SHL9A-6}), with the starboard course change visible in \cref{fig:SHL9A-5}. The highlight here is the model’s predictions in terms of handling grounding hazards in a confined environment. At $t=180s$, the model identifies a higher probability for $C\_COLAV\_M_1$, indicating a collision avoidance maneuver. Even though $p$ and $q$ candidate trajectories could be viable as compliant with the intentions, they have 0\% confidence from the model. Since the model deems that the reference vessel intends to keep a safe distance from the ground ($SDG=true$) and does not intend to sail towards the grounding hazard ($I_G=false$), those trajectories receive zero confidence values. This is evident throughout the rest of the encounter (see \cref{fig:SHL9A-2,fig:SHL9A-3,fig:SHL9A-4}) as well.

In the encounter shown in \cref{fig:G7W2E-1,fig:G7W2E-2,fig:G7W2E-3,fig:G7W2E-4}, a fjord crossing has occurred between two vessels. At the beginning of the encounter (until $t=360s$), it can be considered both a navigation maneuver and a collision avoidance maneuver, which is reflected in the probabilities of $C\_NAV\_M_1$ and $C\_COLAV\_M_1$ (see \cref{fig:G7W2E-6}). However, with the course change towards the next waypoint between $t=300-480s$ (see \cref{fig:G7W2E-5}) and the closing of the gap between the vessels, the model deems that it is no longer a collision avoidance maneuver but solely a navigation maneuver towards the next waypoint. This is indicated by the probability of $C\_COLAV\_M_1$ dropping to zero at $t=420s$.Without grounding hazards, all four candidate trajectories resulting from starboard maneuvers receive similar confidence percentages up until the probability of $C\_COLAV\_M_1$ goes to zero.

\section{Discussion and Future Work} \label{sec:conclusion}
This paper enhances the existing DBN model \cite{rothmund_intention_2022} for target ship intention inference by integrating grounding hazard and waypoint information. To ensure realism, the model's parameters were derived from a historical AIS dataset. The proposed method was tested and validated using real-world encounters from the same dataset.

It was concluded that when extracting prior probability distributions for the intention nodes, external factors contributing to the intentions need to be investigated in a separate study. For example, what a reference vessel considers a safe distance ($I_{SD}$) may depend on multiple factors, including ship size, speed, and ship type (maneuverability). Therefore, a more realistic estimation of the prior probability distributions may involve using several probability distributions conditional on the respective external factors for that node, as suggested in \cite{rothmund_intention_2022}. The idea in Section \ref{sec:parameter_extraction} is to propose an extraction method such that, if the encounters in the dataset are classified according to these conditions, multiple conditional probability distributions can be derived by applying the same methodology to the classified data.

Two methods have been proposed for utilizing the DBN in a collision avoidance algorithm \cite{rothmund_intention_2022}. The first method involves using the intention probabilities directly within a collision avoidance algorithm that accounts for probabilistic factors, such as Probabilistic Scenario-Based Model Predictive Control (PSB-MPC) \cite{tengesdal_collision_2020}, as demonstrated in \cite{tengesdal_obstacle_2024}. The second method, which is used in this paper to obtain results, involves generating a candidate trajectory that represents a probable future path. The DBN itself is then used to evaluate the compatibility of the suggested trajectory with the predicted intentions.

In this paper, a simple line-of-sight guidance controller is used to generate such candidate trajectories. A nominal path—a straight line from the ship's observed position in its current direction—is first generated. By applying constant course offsets to the guidance controller outputs, multiple trajectories are created that initially deviate from the nominal path before aligning parallel to it, as shown in \cref{fig:29ANT,fig:SHL9A,fig:G7W2E}. However, a more systematic methodology can be employed to generate these trajectories using different trajectory generation techniques, such as Rapidly Exploring Random Tree (RRT), as hinted in \cite{tengesdal_comparative_2025}.

\section*{Acknowledgement}
We thank the Norwegian Coastal Administration for access to AIS data.

\appendix
\section{} \label{sec:app1}
The prior probability distributions of the intention nodes and measurement nodes used to obtain results for this paper are summarized in Table \ref{tab:intention_node_priors} and Table \ref{tab:measurement_node_priors}.
These distributions are derived from a specific historical AIS dataset, and are not universal or typical. The extraction process follows algorithms explained in Section \ref{sec:parameter_extraction}. 
\begin{table}[htb]
\centering
\caption{The initial probability distributions for the intention nodes. $\mathcal{N}(\mu, \sigma)_{[a, b]}$ represent a truncated normal distribution between $a$ and $b$, with a mean value of $\mu$ and a standard deviation of $\sigma$.}
\begin{tabularx}{\columnwidth}{|l|X|}
\hline
        Symbol  & Prior Probability Distribution \\ \hline
        $I_{AT}$ & $\mathcal{N}(2527s, 1120s)_{[0, 5000]}$ \\
        $I_{CC}$ & $[false=0.02, true=0.98]$ \\ 
        $I_{GS}$ & $[false=0.01, true=0.99]$ \\ 
        $I_{P_i}$ & $[higher=0.05, similar=0.90,lower=0.05]$ \\
        $I_{CS_i}$ & $\mathcal{M}_{CS_i}$ (see Table \ref{tab:measurement_nodes}) \\
        $I_{G}$ & $[false=0.99, true=0.01]$ \\
        $I_{SDGS}$ & $\mathcal{N}(436m, 124m)_{[0, 700]}$ \\
        $I_{SDGF}$ & $\mathcal{N}(535m, 120m)_{[0, 800]}$ \\
        $I_{SD}$ & $\mathcal{N}(808m, 430m)_{[0, 1500]}$ \\
        $I_{SDF}$ & $\mathcal{N}(1411m, 472m)_{[0, 2000]}$ \\
        $I_{SDM}$ & $\mathcal{N}(249m, 148m)_{[0, 600]}$ \\
        $I_{U}$ & $[false=0.99, true=0.01]$ \\ 
\hline
\end{tabularx}
\label{tab:intention_node_priors}
\end{table}
\begin{table}[htb]
\centering
\caption{The initial probability distributions for the measurement nodes. $\mathcal{U}_{[a, b]}$ represent a uniform distribution between $a$ and $b$.}
\begin{tabularx}{\columnwidth}{|l|X|}
\hline
        Symbol  & Prior \\ \hline
        $\mathcal{M}_{DCPA_i}$ & $\mathcal{U}_{[0, 1500]}$ \\
        $\mathcal{M}_{DF_i}$ & $\mathcal{U}_{[0, 2000]}$ \\ 
        $\mathcal{M}_{DM_i}$ & $\mathcal{U}_{[0, 600]}$ \\ 
        $\mathcal{M}_{P_i}$ & $[false=0.5, true=0.5]$ \\
        $\mathcal{M}_{PS_i}$ & $[starboard=0.5, port=0.5]$\\
        $\mathcal{M}_{MPS_i}$ & $[starboard=0.5, port=0.5]$\\
        $\mathcal{M}_{TCPA_i}$ & $\mathcal{U}_{[0, 5000]}$ \\
        $\mathcal{M}_{CIC}$ & $[starboard=0.33, port=0.33, straight=0.33]$ \\
        $\mathcal{M}_{CIS}$ & $[higher=0.33, lower=0.33, none=0.33]$ \\
        $\mathcal{M}_{CCC}$ & $[false=0.5, true=0.5]$ \\
        $\mathcal{M}_{CS_i}$ & $[HO=0.2, OT\_en=0.2, OT\_ing=0.2, CR\_PS=0.2, CR\_SS=0.2]$ \\
        $\mathcal{M}_{DGSB}$ & $\mathcal{U}_{[0, 700]}$ \\
        $\mathcal{M}_{DGPS}$ & $\mathcal{U}_{[0, 700]}$ \\
        $\mathcal{M}_{DGF}$ & $\mathcal{U}_{[0, 800]}$ \\ 
        $\mathcal{M}_{WPRB}$ & $[decreasing=0.33, increasing=0.33, neither=0.33]$ \\ 
        $\mathcal{M}_{WPRD}$ & $[decreasing=0.33, increasing=0.33, neither=0.33]$ \\ 
        $\mathcal{M}_{WPAH}$ & $[false=0.5, true=0.5]$ \\ 
\hline
\end{tabularx}
\label{tab:measurement_node_priors}
\end{table}

%%       \citet{<label>} ==> Lamport (1994)
%%       \citep{<label>} ==> (Lamport, 1994)
%%

\bibliographystyle{elsarticle-harv} 
\bibliography{references}





% \bibitem[Lamport(1994)]{lamport94}
%   \textit{\LaTeX: a document preparation system},
%   Addison Wesley, Massachusetts,

% \end{thebibliography}
\end{document}